\documentclass[journal]{IEEEtran}
\usepackage{graphicx}
\usepackage{multirow}
\usepackage{amsmath,amssymb} 
\usepackage{epstopdf}
\usepackage{algorithm}
\usepackage{algorithmic}
\usepackage{subfigure}
\usepackage{bm}
\begin{document}
%
\title{Skeleton-aided Articulated Motion Generation}
%
%
%

\author{Yichao Yan, Jingwei Xu, Bingbing Ni, Xiaokang Yang}

\maketitle

\begin{abstract}
This work makes the first attempt to generate articulated human motion sequence from a single image. On one hand, we utilize paired inputs including human skeleton information as motion embedding and a single human image as appearance reference, to generate novel motion frames based on the conditional GAN infrastructure. On the other hand, a triplet loss is employed to pursue appearance smoothness between consecutive frames. As the proposed framework is capable of jointly exploiting the image appearance space and articulated/kinematic motion space, it generates realistic articulated motion sequence, in contrast to most previous video generation methods which yield blurred motion effects. We test our model on two human action datasets including KTH and Human3.6M, and the proposed framework generates very promising results on both datasets.
\end{abstract}


\section{Introduction}
Object motion prediction and generation in the videos is a key factor in video analysis, and it has potential application to smart surveillance, human-computer interaction and other applications.
Generative models such as GAN~\cite{goodfellow2014generative} have achieved great success on image generation, but how to generate videos with motion dynamics is rarely explored.
Although recent developments of convolutional neural network (CNN) and recurrent neural network (RNN/LSTM~\cite{DBLP:journals/neco/HochreiterS97}) have made great success on action classification task~\cite{DBLP:conf/cvpr/KarpathyTSLSF14,DBLP:conf/mm/WuWJYX15,DBLP:journals/pami/DonahueHRVGSD17}, motion generation is still challenging because it often involves high-dimensional data with complex temporal dynamics.
In particular, previous video generation methods~\cite{DBLP:conf/icml/SrivastavaMS15,finn2016unsupervised,DBLP:conf/nips/VondrickPT16,DBLP:conf/nips/OhGLLS15} are only good at simulating rigid movement of objects. In the case of articulated movement (e.g., human motion), these methods mostly yield blurred effects for various body parts.

\begin{figure}
\includegraphics[width=3.5in]{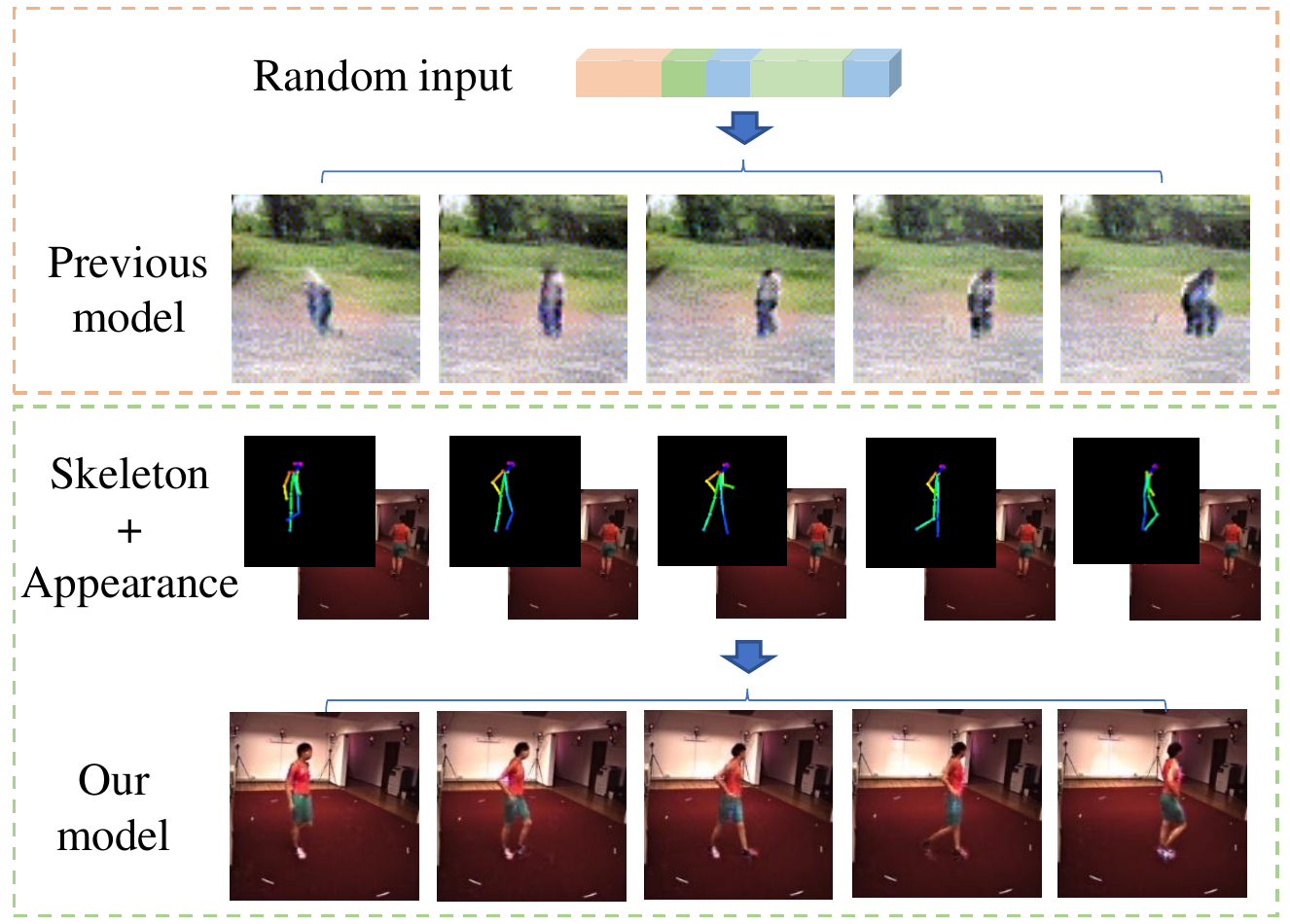}
\caption{Overview of the proposed framework. Previous video generation methods do not explicitly model the articulated structure of the foreground motion objects, the generated videos usually suffer from great deformation (see the top part). The bottom part gives an illustration of our approach, and we utilize the appearance information as well as the skeleton information to generate motion sequence.  }
\end{figure}

Existing video generation methods mainly focus on two tasks.
The first one is video prediction~\cite{DBLP:conf/nips/OhGLLS15,DBLP:conf/icml/SrivastavaMS15,DBLP:journals/corr/LotterKC16,DBLP:journals/corr/MathieuCL15,DBLP:journals/corr/AmersfoortKRSTC17,DBLP:journals/corr/RanzatoSBMCC14}, i.e., the models need to learn the motion patterns from a sequence of observed frames and to predict/generate the next frames. These methods are usually based on a recurrent structure (RNN of LSTM), despite of the good ability of the RNN/LSTM to model sequential data, they usually achieve good results only for short-term predictions where the videos are simple and quiet predictable. While the long-term prediction results usually suffer from low image quality, such as blur and object deformation.
The second type of methods aim to directly generate a sequence of frames based on a single input~\cite{DBLP:conf/eccv/WalkerDGH16} or only the scene types~\cite{DBLP:conf/nips/VondrickPT16}.
This task is more challenging as the motion patterns can no longer be observed during test phase.
These methods employ the GAN model to generate the spatio-temporal cuboids or employ the Variational Autoencoders~\cite{DBLP:journals/corr/KingmaW13} to forecast the dense trajectory of pixels in the scene.
However, the objects in the scene can move arbitrarily if no geometric constrains are given to foreground objects, which will result in great deformation on the generated objects.
Furthermore, we observe a shared limitation over these two type of methods, i.e., the articulated structures of the foreground motion objects (i.e., human) are not well modeled in the generation model. As previous generative methods only take the whole appearance as input, it will be difficult for the model to learn the structural relationship among the articulations/parts if given no supervision, thus resulting in great deformation during motion.
Limited by this constraint, the quality of the generated videos are far from satisfaction.

In this work, we propose to use skeleton information to help generate articulated motion, which is motivated by the following observations.
On the one hand, the articulated motion is usually under strong structure/geometric constrain, which can be well represented by the skeletons.
On the other hand, compared to images with high dimension, the skeleton (coordinates of body parts) serves as a very good low dimensional embedding for human motion. Therefore it could be used as underlying status parameters to generate flexible poses. Also, skeletons can be mapped to image one-by-one, this avoids the long-term prediction problem shared by previous methods.
Moreover, recent development on human pose estimation techniques has made skeleton data easy to access, thus avoiding heavy human annotations.

Our problem is defined as follows. Suppose we have a sequence of skeletons and a single image of human appearance, the task is to generate a sequence of articulated motion images. This task can be further decomposed into two sub-problems: 1) how to generate realistic articulated motion (i.e., instead of blobs); and 2) how to generate appearance, which is adapted to every generated image frame.
On one hand, motion generation (i.e., generating different human pose) is addressed by a GAN-like network. Recently, GAN has achieved great success in image generation, domain adaption and most importantly, image-to-image translation. The motion generation process can be naturally transformed into skeleton-to-image translation problem, which can be naturally handled by a conditional GAN model (i.e., in this work, we employ a GAN loss and an L1 loss to ensure smooth image-to-image translation). On the other hand, if given no appearance information (cloth color, bodily form), appearance of the generated image sequences cannot be controlled, i.e., the generated appearances might differ from image to image. This violates the rule that the appearance of an object should be consistent during the entire motion sequence. To address this issue, we choose to generate the motion sequence based on both skeleton sequence and an appearance image, which is realized with a specially designed generator. Furthermore, in order to ensure inter frame continuity, and we also employ a triplet loss which aims to penalize the generation loss if the adjacent frames have larger distance than the non-adjacent frames.
\begin{figure*}
\includegraphics[width=7in]{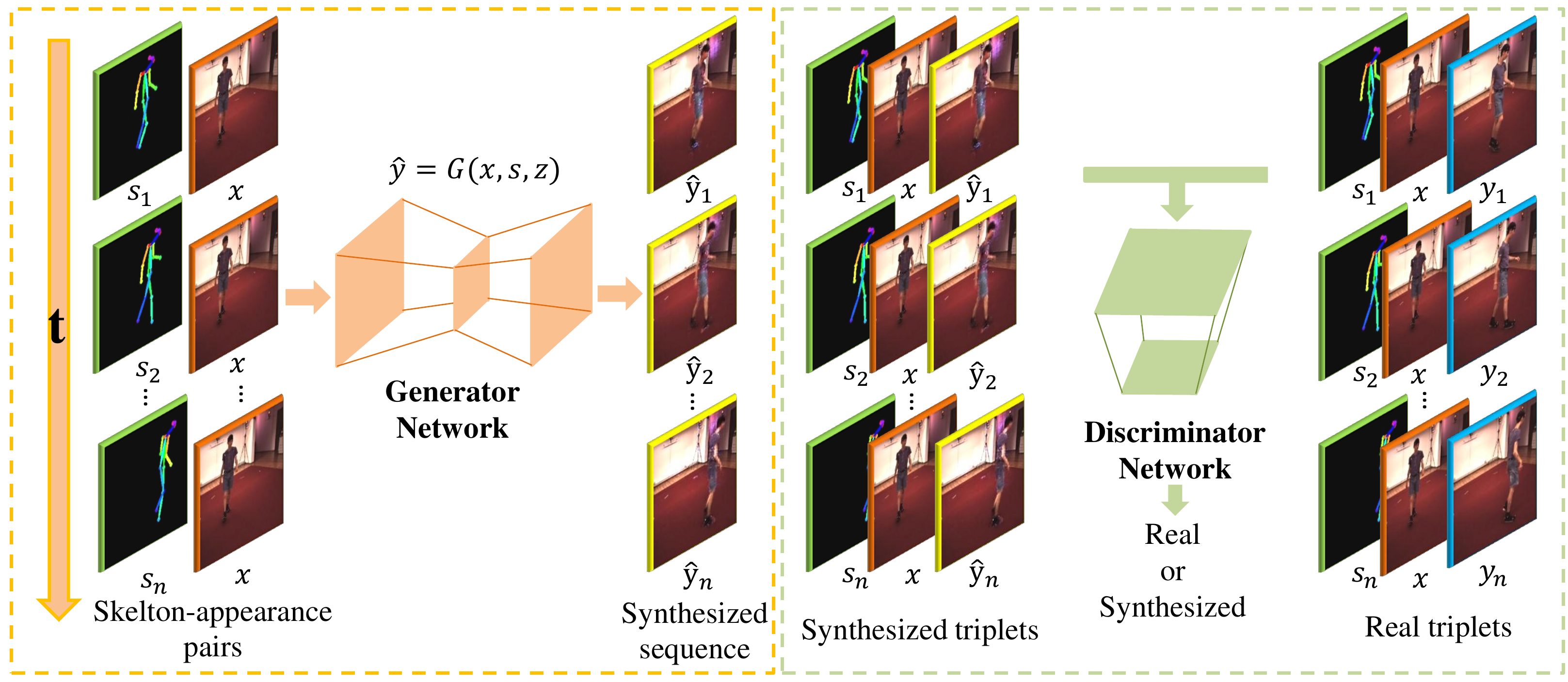}
\caption{Architecture of the generation and discrimination network. The inputs for the generator are the skeleton-appearance pairs $(s,x)$ and generate the synthesized sequence $\hat{Y}=\{\hat{y}_1,...,\hat{y}_n\}$. The discriminator D tries to distinguish real triplets $(x,s,y)$ and synthesized triplets $(x,s,\hat{y})$.}
\label{fig:arc}
\end{figure*}

This work is among the very few works for complete video generation. To the best of our knowledge, this is the first work to employ skeleton information to help generate videos. Our key contribution is that the proposed method can generate images/videos with large scale geometry change, where previous methods achieve little success. We test our model on two human action datasets including KTH and Human3.6M, and the proposed framework generates very promising results on both datasets.

\section{Related work}
\textbf{Human Motion Analysis}. In computer vision, human motion analysis is a broad concept which focuses on the understanding and applications of human motion patterns. It has been receiving increasing attention for a long time and has been applied to enormous applications, such as content-based video retrieval, visual surveillance, and man-computer interfaces~\cite{poppe2007vision,aggarwal1997human,ji2010advances}. Among the researches of human motion analysis, variant human body models (e.g., stick figure model~\cite{ju1996cardboard,howe1999bayesian}, cardboard model~\cite{rashid1980towards}, $3D$ volumetric model~\cite{kehl2006markerless}) play an important role. These models involve low-level processes on human body structures and cover the kinematic properties of the body, which build the foundation in solving different problems, including human motion tracking~\cite{huang2002model,ong2006viewpoint,mikic2003human}, action recognition~\cite{du2015hierarchical,papadopoulos2014real,grushin2013robust,gong2014structured}, and pose estimation~\cite{andriluka2009pictorial,andriluka20142d,toshev2014deeppose}. Motivated by these successful applications based on well-defined human body models, in this paper, we pay attention to another more challenging task that attempts to generate consistent human motion sequences based on the correspondent skeleton and appearance information.

\textbf{Image Generation}. Early works for image generation usually make efforts on simple texture synthesis with hand-crafted features~\cite{portilla2000parametric}.
During the past few years, two generation models have been attracting more and more attention, i.e., the variational autoencoder (VAE)~\cite{DBLP:journals/corr/KingmaW13} and the generative adversarial network (GAN)~\cite{goodfellow2014generative}.
VAE is a classical method which aims to model complicated distribution and it has been widely applied in various generative tasks. Gregor et al.~\cite{gregor2015draw} propose a sequential generative model which extends the original VAE with recurrent neural networks and attention mechanism. Another interesting model is proposed by Yan et al.~\cite{yan2016attribute2image}, they develop a layered generative model based on conditional VAE.
GAN is also a popular generative model and many recent works are built on it. Some works improve the architecture of original GAN for better performances~\cite{zhao2016energy,chen2016infogan,qi2017loss,arjovsky2017wasserstein}. Conditional generative adversarial network (CGAN)~\cite{mirza2014conditional} gives extra information to the input as condition, and the output is constrained by the input conditions. CGAN has been further extended by~\cite{DBLP:journals/corr/IsolaZZE16,zhu2017unpaired,pmlr-v70-kim17a} to solve the image-to-image translation problem, which gives inspiration of our model proposed in this paper. Other applications such as image super-resolution~\cite{ledig2016photo}, image edition~\cite{brock2016neural} and unsupervised representation learning~\cite{radford2015unsupervised} also show impressive results.

\textbf{Video Generation}. Our problem is closely related to video generation or prediction.
Video texture based methods~\cite{schodl2000video,agarwala2005panoramic,liao2013automated} can generate periodic motion sequences if an input reference video is given.
Lotter et al.~\cite{DBLP:journals/corr/LotterKC16} propose a predictive neural network motivated by the concepts from neuroscience.
Finn et al.~\cite{finn2016unsupervised} develop an action-conditioned video prediction model which concentrates on pixel motion and Mathieu et al.~\cite{DBLP:journals/corr/MathieuCL15} introduce multi-scale architecture to reduce the deformation in the predictions. Instead of focusing on pixel level prediction, Van et al.~\cite{van2017transformation} attempt to predict the transformations between frames. Some works also utilize GAN or VAE in video generation. Vondrick et al.~\cite{DBLP:conf/nips/VondrickPT16} propose a GAN model which generates static background and dynamic foreground sequences separately. Xue et al.~\cite{xue2016visual} introduce conditional VAE and build a cross convolutional network which encodes image and motion information for generation. Although variant models are proposed, the results of these methods are usually limited by two issues: 1) the deformation of the foreground object is serious, 2) and the inter-frame consistency cannot be well maintained. These problems inspire us that in order to generate more realistic videos, strong motion constrains are needed during the generation process. Therefore, we employ skeleton information to guide our model for motion generation.


\section{Method}
Our problem is defined as follows. Given an image $x$ containing the foreground person (the appearance reference image), we would like to generate a sequence of images $Y=\{y_1,..., y_n\}$ that share the same appearance, and the foreground objects should keep a specific motion pattern as well (e.g., walking, running). In other words, we would like to generate articulated motion from a single static image.
This is challenging because a person could have infinite move patterns. The first step is to choose a specific move pattern for the sequence. As skeleton can well represent a person's motion, we employ a sequence of skeletons $S=\{s_1,..., s_n\}$ as prior knowledge for the motion. The remaining problem is to generate the output sequence based on the appearance image and skeleton sequence: $\{x,S\} \rightarrow Y$. Here, we propose a skeleton conditioned GAN to model this mapping.
The second problem is to maintain appearance-smoothness between consecutive frames, we employ a triplet loss on the generator for this purpose.
The details of the proposed method are given in the following of this section.

\subsection{Skeleton Conditioned GAN}
Different from the previous image-to-image translation model, where only a single input image is mapped to the output, our generative model is conditioned on two inputs, i.e., the appearance reference image $x$ and the skeleton image $s$.
The value function of the Conditional GAN (CGAN) model is expressed as follows:
\begin{equation}
\begin{split}
\mathcal{L}_{c}(G,D)&=\mathbb{E}_{x,s,y\sim p_{data}(x,s,y)}[\log D(x,s,y)]\\
&+\mathbb{E}_{x,s\sim p_{data}(x,s),z\sim p_{z}(z)}[\log(1-D(x,s,G(x,s,z)))],
\end{split}
\label{func:sub}
\end{equation}
where the generator $G$ tries to produce a new frame, and a discriminator $D$ tries to distinguish real triplets $(x,s,y)$ and synthesized triplets $(x,s,G(x,s,z))$.
The architectures of the generator and discriminator model are illustrated in Figure~\ref{fig:arc}.

Previous methods~\cite{DBLP:journals/corr/IsolaZZE16,DBLP:journals/corr/MathieuCL15} observe that using the CGAN loss alone will give sharp results with artifacts, and it's beneficial adding a contractive loss such as L1 loss. Although this may cause blur effect, mixing these two losses will generate overall better results. The L1 loss can be expressed as:
\begin{equation}
\mathcal{L}_{L_1}(G)=\mathbb{E}_{x,s\sim p_{data}(x,s),z\sim p_{z}(z)}[\|y-G(x,s,z)\|_1].
\end{equation}

\begin{figure}[t]
\center
\includegraphics[width=3.3in]{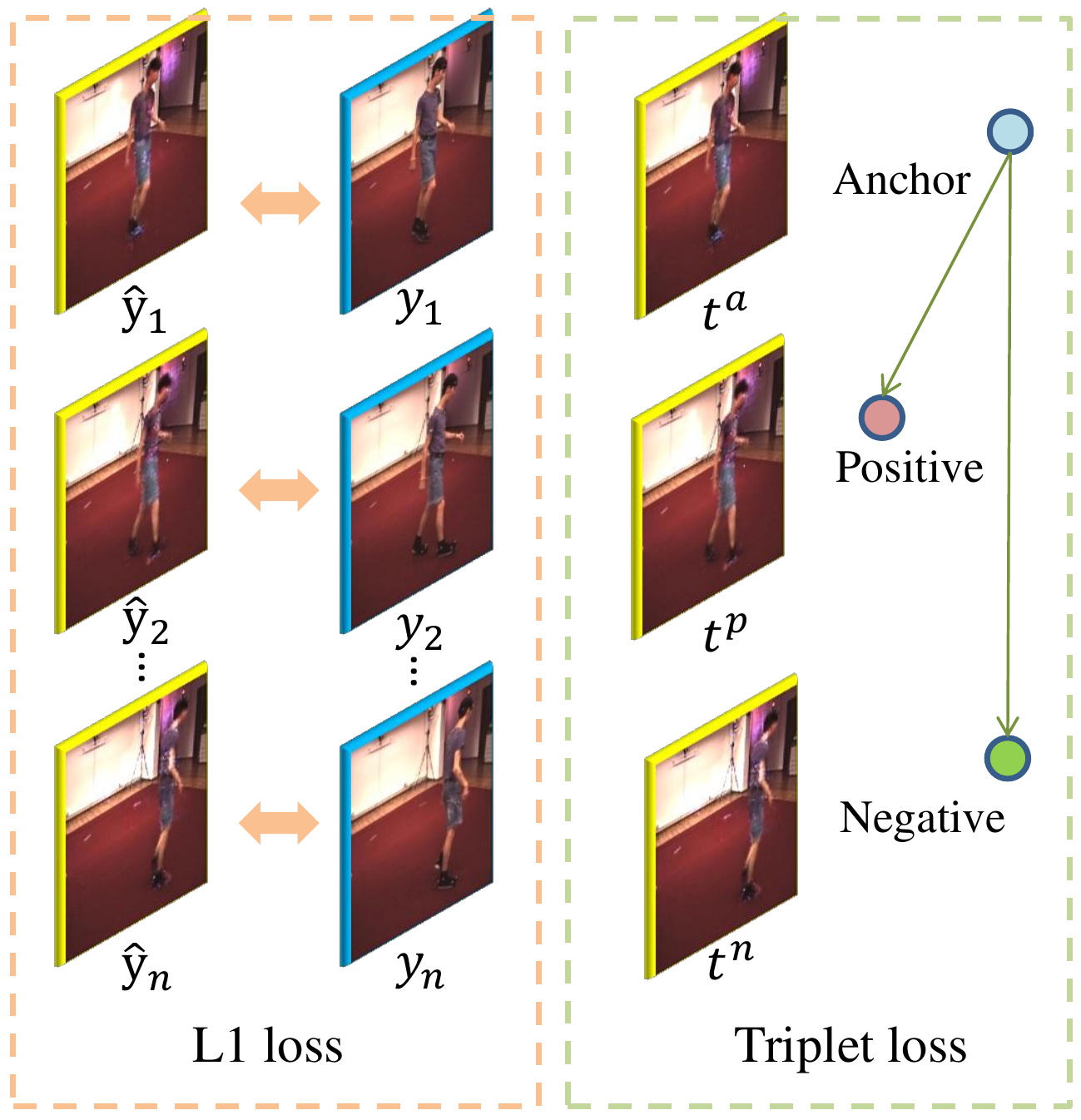}
\caption{Loss terms of the model. Despite of the GAN loss, we take two additional loss terms, i.e., the L1 loss to enhance the image-to-image translation quality, and the triplet loss to guarantee the continuity of the generated motion sequence.}
\label{fig:losses}
\end{figure}

Notice that our goal is to generate continuous motions rather than individual images. Thus it is important to consider the motion continuity and appearance smoothness of the adjacent frames. Although this can be achieved by training a perfect generator that precisely maps the input appearance image into a new pose specified by the input skeleton, the perfect generator cannot be achieved in practice because a single appearance image does not contain the complete information of the moving object. e.g., for an image of a person, some parts are inevitably occluded, which can't be generated perfectly through training. To address this issue, we propose to utilize a triplet loss that motivates adjacent frames to be more similar than the far-away frames. First, we need to construct the triplet set $\mathcal{T} = \{t^a_i, t^p_i, t^n_i\}_{i=1}^{m}$ from the generated samples: $\hat{Y}=\{\hat{y}_1,...,\hat{y}_n\}$,
where $\hat{y}_j = G(x, s_j, z), 1\le j \le n$, and $m$ is the number of triplets selected for training. The anchor of the triplet can be randomly chosen from the generated samples, say $t^a_i= \hat{y}_j$, the positive sample can select a sample adjacent to the anchor (e.g., $t^p_i= \hat{y}_{j+1}$), and the negative sample can choose a far-away sample (e.g., $t^n_i= \hat{y}_{j+5}$). We would like that the distance between anchor and positive is smaller than that of anchor and negative, thus the loss function can be expressed as:
\begin{equation}
\mathcal{L}_{tri}(G) = \sum_{i=1}^{m}[\|t^a_i-t^p_i\|_2^2-\|t^a_i-t^n_i\|_2^2 + \alpha]_{+}.
\end{equation}
In our experiments, we also tried to replace the L2 norm with L1 norm in the triplet loss, but we don't observe performance gain. The loss terms are illustrated in Figure~\ref{fig:losses}.

The overall objective function is:
\begin{equation}
\mathcal{L}(G,D)=\mathcal{L}_{c}(G,D)+\lambda\mathcal{L}_{L_1}(G)+\beta\mathcal{L}_{tri}(G),
\label{eq:l}
\end{equation}
where $\lambda$ and $\beta$ are the weights for different loss terms.
We aim to solve
\begin{equation}
G^{*}=\arg\min \limits_{G}\max \limits_{D}\mathcal{L}(G,D).
\end{equation}

\subsection{Generator architecture}
The basic structure of our network is built upon~\cite{DBLP:journals/corr/IsolaZZE16}, in which the generator usually follows the encoder-decoder structure. Specifically, the U-net structure~\cite{DBLP:conf/miccai/RonnebergerFB15} has proven to be more effective for image-to-image translation tasks, because it adds skip connections between encoder and decoder. Such a structure enables the information to share between the inputs and outputs, and thus achieving success for tasks such as image colorization, image segmentation, etc.
We also adopt this structure for the generator. In our case, the generator needs to translate two inputs (i.e., the appearance reference image $x$ and the skeleton image $s$) into a single output $y$. Therefore, the encoder takes two inputs simultaneously and shuttles the information to the decoder.
\begin{figure}
\centering
\subfigure[Stacked Generator.]{
\label{fig:stacked_encoder}
\includegraphics[width=1.53in]{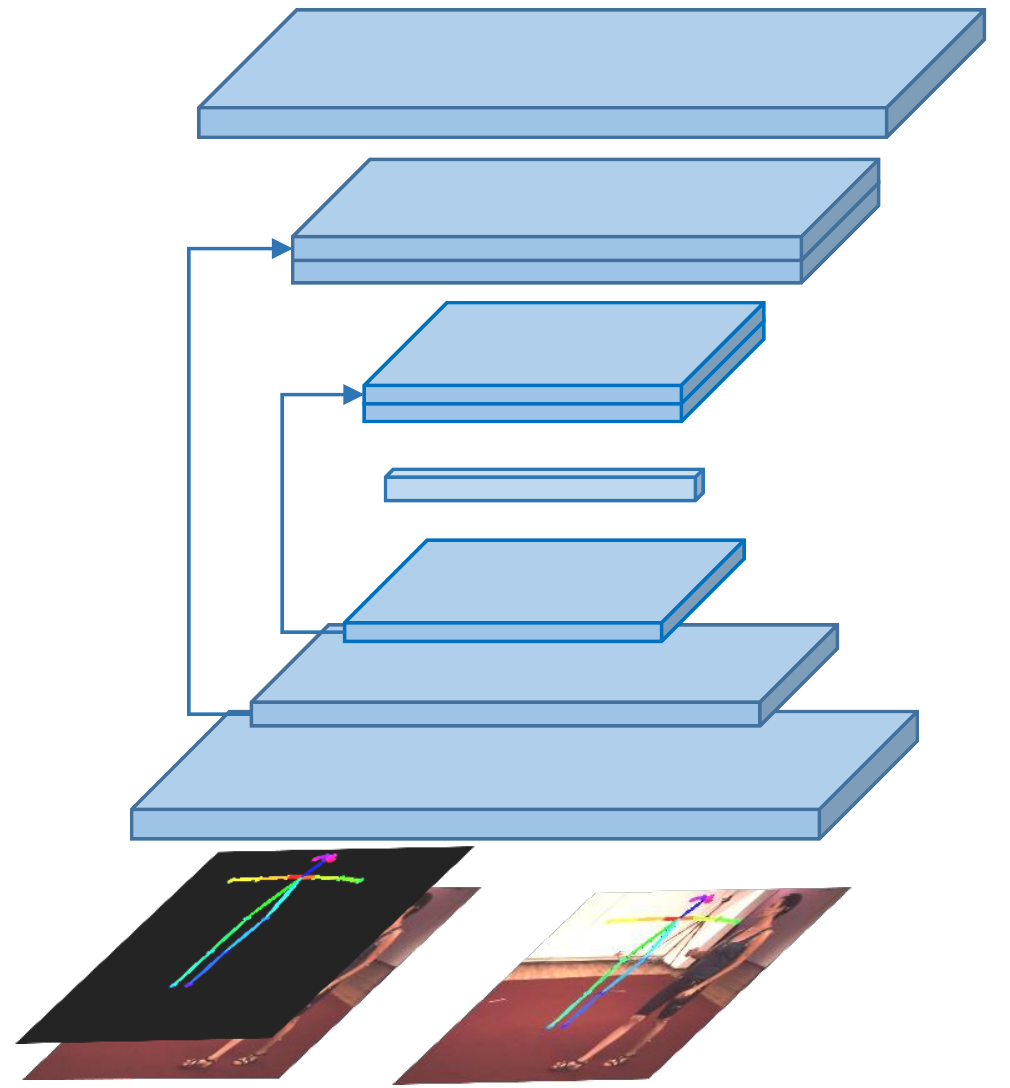}
}
\hspace{.01in}
\subfigure[Siamese Generator.]{
\label{fig:siamese_encoder}
\includegraphics[width=1.53in]{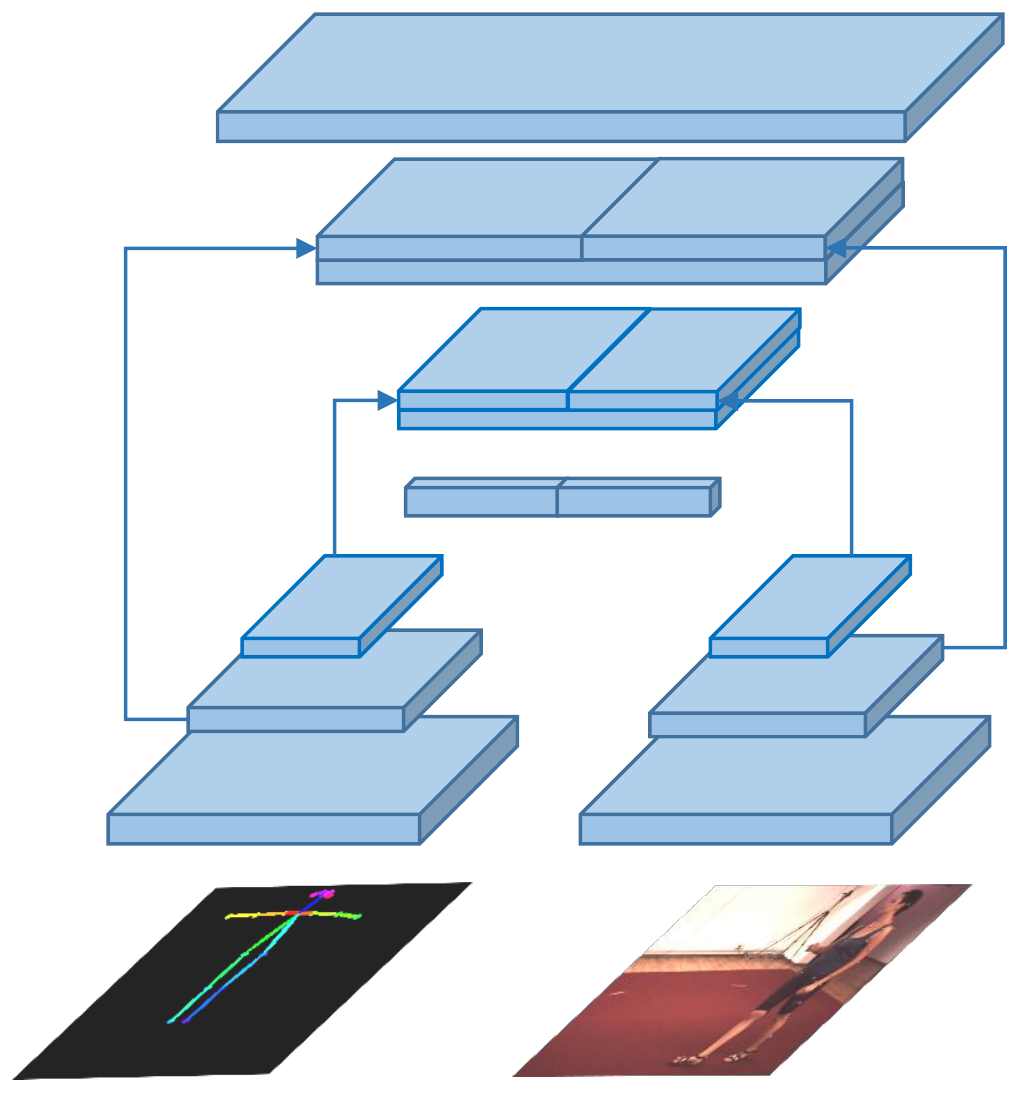}
}
\caption{Structures of the generator. For both the stacked generator and the Siamese generator, we take the U-Net structure for both generators. }
\label{fig:encoder}
\end{figure}

\textbf{Stacked Generator}. There are multiple options to design the structure of the encoder. The most straightforward way is to stack the two input images as a single input (i.e., $i = [x,s]$), thus the standard encoding network can be applied directly. We denote this structure as \textbf{stacked generator 1}. We further observe that the skeleton images have completely black ground which is not informative at all, and all the motion information is contained in the pose of the skeleton. Therefore, we can directly draw the skeleton on the appearance image instead. The resulted image sequence can be viewed as a skeleton moving on the appearance image. In this case, the inputs for the encoder are standard images, and have no distinction with the traditional image-to-image task. We denote this structure as \textbf{stacked generator 2}. The two stacked structures are illustrated in Figure~\ref{fig:stacked_encoder}.

\begin{table}[]
\centering
\caption{Detailed structure of the Generator.}
\label{generator}
\begin{tabular}{|c|c|}
\hline
\multicolumn{2}{|c|}{\textbf{Encoder}}                \\ \hline
Layer & Input size: 256$\times$256$\times$6(3)                       \\ \hline
1     & Conv-(64,K4$\times$4,S2), lReLU-(0.2)                   \\
2     & Conv-(128,K4$\times$4,S2), BN, lReLU-(0.2)                  \\
3     & Conv-(256,K4$\times$4,S2), BN, lReLU-(0.2)                  \\
4-8   & Conv-(512,K4$\times$4,S2), BN, lReLU-(0.2)                  \\ \hline
\multicolumn{2}{|c|}{\textbf{Decoder}}                \\ \hline
Layer & Input size: 60$\times$60$\times$1(2)                            \\ \hline
1     & FConv-(512,K4$\times$4,S2), BN, Dropout-(0.5), ReLU  \\
2-5   & FConv-(1024,K4$\times$4,S2), BN, Dropout-(0.5), ReLU \\
6     & FConv-(512,K4$\times$4,S2), BN, Dropout-(0.5), ReLU  \\
7     & FConv-(256,K4$\times$4,S2), BN, Dropout-(0.5), ReLU  \\
8     & FConv-(128,K4$\times$4,S2), BN, Dropout-(0.5), ReLU  \\ \hline
\end{tabular}
\end{table}

The stacked encoders provide a simple solution for simultaneously encoding two input images by stacking the two input as an ensemble. However, the two inputs usually contain different information, which needs to be modeled separately. To this end, we design a second structure.

\textbf{Siamese Generator}. In particular, each input image can be modeled by an encoding network, and the features are concatenated at the bottleneck layer for the decoder. In this way, the different information of both the appearance image and the skeleton image can be well modeled respectively.
We denote this structure as the \textbf{Siamese generator}, because it has Siamese structure. See Figure~\ref{fig:siamese_encoder} for an illustration.

The detailed structure of the generator is illustrated in Table~\ref{generator}, where the encoder is composed of convolution (Conv), batch normalization (BN) layers and leaky rectified linear unit (lReLU) layers, and the decoder is composed of fractional length convolutional (FConv) layers, BN layer, Dropout layer and the ReLU layers.
For gray-scale inputs, we replicate their channel 3 times so that the network doesn't distinguish RGB inputs and gray-scale inputs.
We resize all the input images into a fix size, i.e., 256$\times$256$\times$3. Therefore, the input size for the stacked generators is 256$\times$256$\times$6 because they stack two inputs. And for the Siamese generator, the number of channel is 3 all the inputs.  Also notice that each of the Siamese structure has exactly the same structure as the encoder, the output of each encoder is concatenated as inputs (60$\times$60$\times$2) for the decoder.

\subsection{Discriminator architecture}
The discriminator needs to be able to classify the realistic triplets $(x,s,y)$ from the synthesized triplets $(x,s,G(x,s,z))$, thus the inputs for the discriminator network are three images. Similar to the encoding network, we can also stack all the three inputs as a 9-dimensional input, or we can extract features from each inputs, and then combine them to make a decision. However, we observe little difference using these two structures during our experiments, therefore we only report the results of the stacked structure.
The discriminator structure is illustrated in Table~\ref{discriminator}.

\begin{table}[]
\centering
\caption{Detailed structure of the Discriminator.}
\label{discriminator}
\begin{tabular}{|c|c|}
\hline
\multicolumn{2}{|c|}{\textbf{Discriminator}}                \\ \hline
Layer & Input size: 256$\times$256$\times$9                       \\ \hline
1     & Conv-(64,K4$\times$4,S2), lReLU-(0.2)                   \\
2     & Conv-(128,K4$\times$4,S2), BN, lReLU-(0.2)                  \\
3     & Conv-(256,K4$\times$4,S2), BN, lReLU-(0.2)                  \\
4-6   & Conv-(512,K4$\times$4,S2), BN, lReLU-(0.2)                  \\      \hline
\end{tabular}
\end{table}

\subsection{Learning and Implementation}
Our implementation is based on a modified version of image-to-image translation network~\cite{DBLP:journals/corr/IsolaZZE16} on Tensorflow~\cite{DBLP:journals/corr/AbadiABBCCCDDDG16}. We report results for several architectures. For all the models, we alternatively train the discriminator and then the generator. We train the generator and discriminator with stochastic gradient descent, with a fixed learning rate as 0.0002 and the Adam optimizer. All the models are trained for 30 epochs. We find that small batch size leads to more appealing results. So the batch size is set as 10 for all the experiments to balance the generation quality and training time. All the videos are re-scaled into range $[-1,1]$ as normalization. The random noise $z$ is not explicitly sampled from Gaussian distribution, it appears only in the form of dropout, which is in consistent with~\cite{DBLP:journals/corr/IsolaZZE16}. The skeletons are extracted using the real-time human pose estimator~\cite{cao2016realtime}. And we also use the same estimator to detect the generated human motion sequence to compare with the ground truth pose estimation. Because we use three kinds of loss function to train the network, we try different weights of each term in training phase to get optimal results. More detailed analysis on loss function is in section 4.3.

\section{Experiments}
In this section, we present extensive experimental evaluations and in-depth analysis of the proposed method. The evaluations are performed on the following two human action datasets:

\textbf{KTH dataset}. This dataset contains several types of human actions in the outdoor environment, and all the videos are gray scale. We experiment on three types of actions, i.e., \textit{walking}, \textit{running} and \textit{hand waving}. And we choose these actions because their motion parts are different, \textit{walking} mainly involve the movement of legs, \textit{hand waving} involve the movement of arms, and \textit{running} involves both the movement of arms and legs. Each type of action contains 100 videos, and we divide the dataset into training set containing 80 videos and test set containing 20 videos.

\textbf{Human3.6M dataset}. This dataset was collected in an indoor environment, there are 4 cameras working simultaneously, i.e., we have access to 4 views of each action. We use the \textit{walking} scenario in this dataset. The dataset also contains foreground segmentations, therefore we experiment on both videos with and without background.

As there exists no standard evaluation protocol for image and video generation tasks, in order to evaluate the quality of the generated videos, we report both qualitative and quantitative results.

\subsection{Structure Analysis}
\begin{figure}
\includegraphics[width=3.5in]{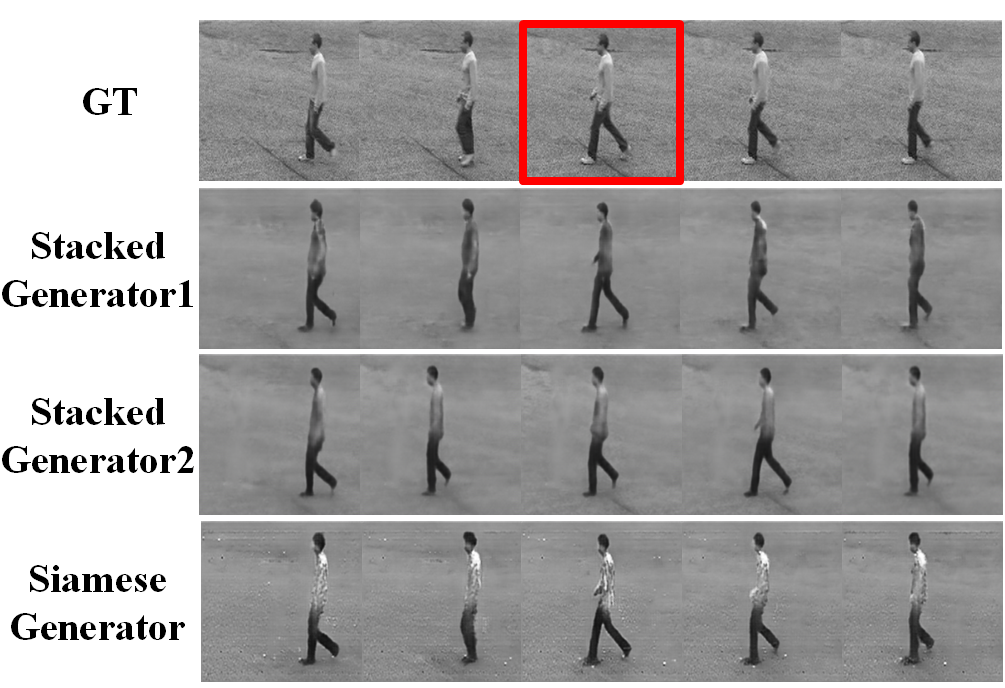}
\caption{Examples generated by different generator structure. The first row contains the ground truth motion sequence, the image with red bounding box is the appearance reference image.}
\label{fig:generator}
\end{figure}
We propose the \textit{stacked structure} and the \textit{Siamese structure} for the generator. We evaluate the performance of the two structure in this part, and the results of different generator structures are compared in Figure~\ref{fig:generator}. The image with red bounding box is the appearance reference image. Benefitting from the U-Net structure, all the generators achieve to generate realistic motion patterns. The two \textit{stacked generators} generate similar results. However, they suffer from a major issue, i.e., the generated sequences do not preserve the appearance of the reference image any more. For example, the subject in the reference image is in white T-shirt and black pants. The results generated by the stacked generators are in dark cloth.
In contrast, the results generated by the \textit{Siamese generator} are more similar to the ground truth.
This is mainly because that the stacked structure does not distinct the pose and appearance information, which is encoded by the same network. However, the appearance images and the skeleton images are under different distributions, encoding them with two separate networks will better fit the two different distributions. Therefore, the \textit{Siamese generator} achieves better results. It not only encodes the motion pattern, but also successfully encodes the appearance information, thus the generated sequence shares high appearance similarity with the reference image.
In the following of this paper, we report the results of the \textit{Siamese generator}.

\subsection{Motion Generation}
\begin{figure*}
\centering
\setlength{\tabcolsep}{0pt}
\renewcommand{\arraystretch}{0}

\subfigure[Walking.]{
\label{fig:subfig:a}
\includegraphics[width=\textwidth]{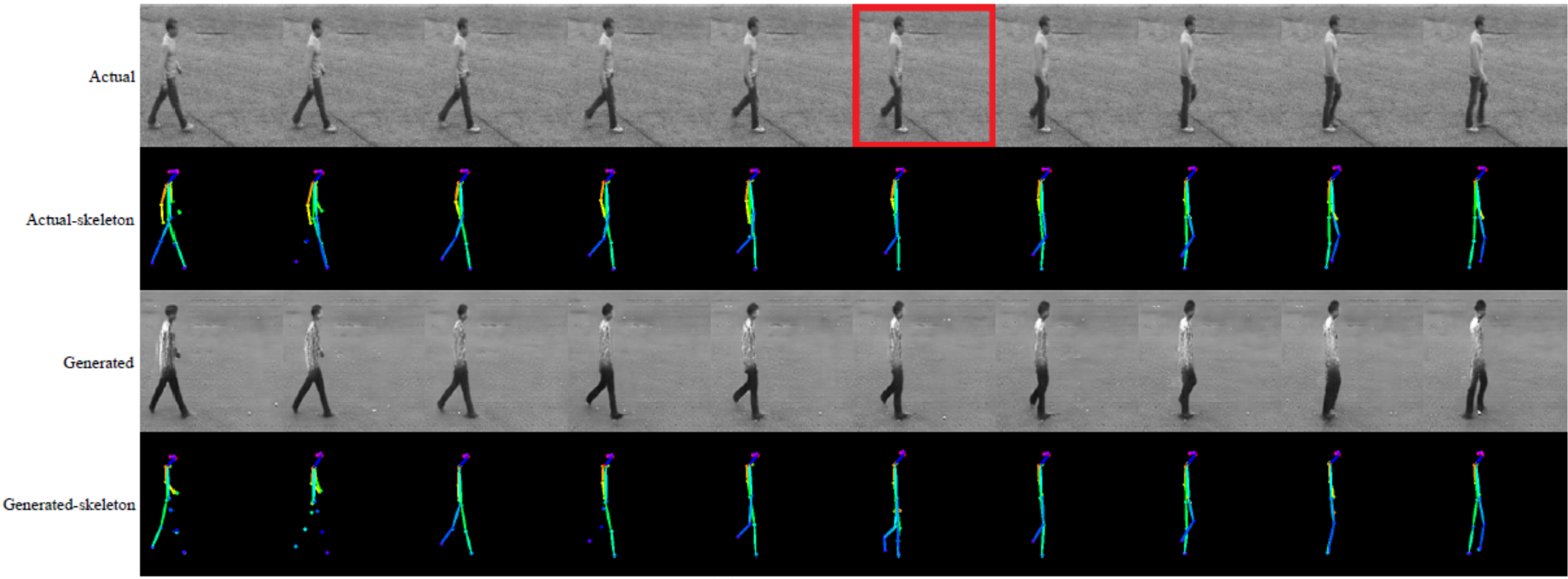}
}

\subfigure[Running.]{
\label{fig:subfig:b}
\includegraphics[width=\textwidth]{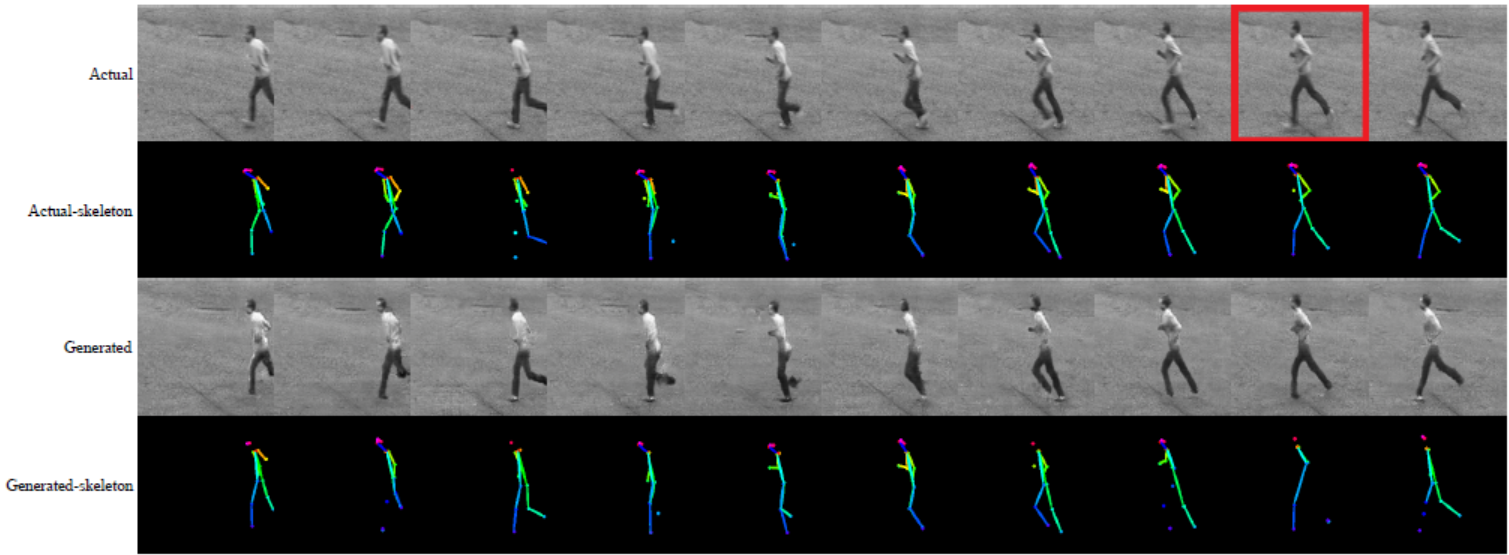}
}

\subfigure[Hand waving.]{
\label{fig:subfig:c}
\includegraphics[width=\textwidth]{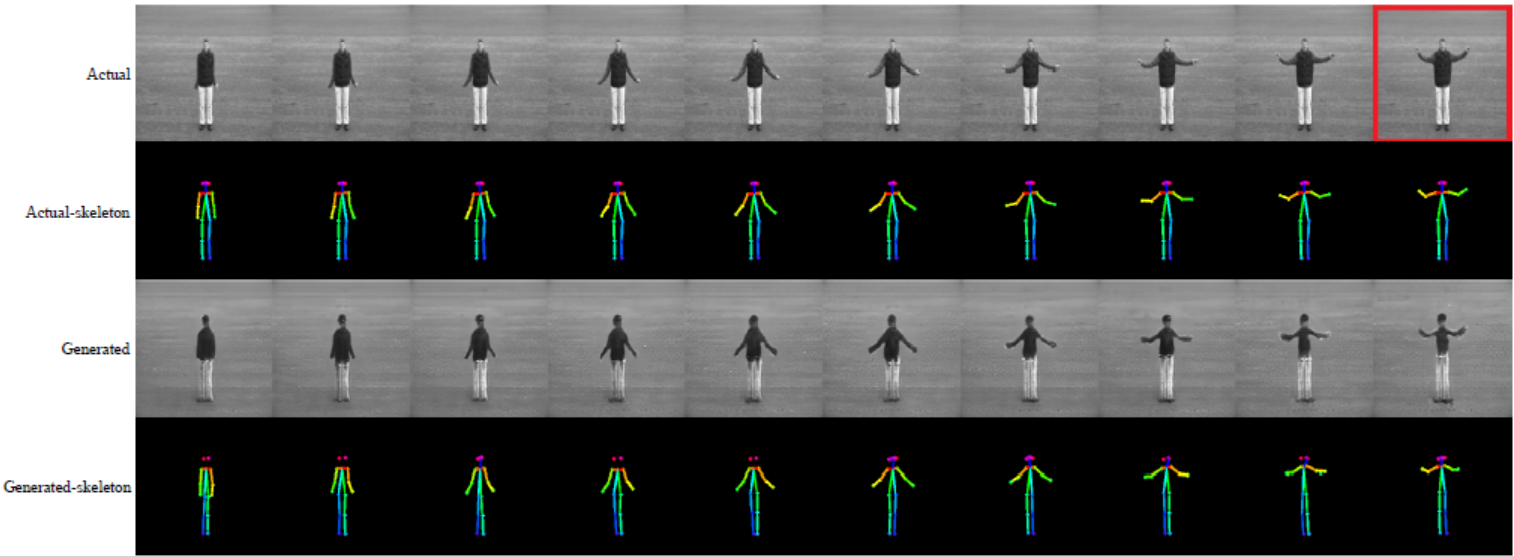}
}
\caption{Generated samples on KTH dataset. The first rows contain the ground truth motion sequences, and the images with red bounding boxes are the appearance reference images. The second rows are the ground truth reference skeletons. The third rows are the samples generated by our model, and the last rows are the detected skeletons of the generated sequences.}
\label{fig:kth}
\end{figure*}

Some generated sequences on KTH dataset are visualized in Figure~\ref{fig:kth}. For each example, the first row is the ground-truth sequence, and the second row represents the detected skeleton sequence. The generated results are given in the third row. We re-run the skeleton detector on the generated sequence, and the resulted skeleton sequence is shown in the last row.
We have three observations: 1) the foreground subjects are naturally generated in the scene. Different from~\cite{DBLP:conf/nips/VondrickPT16}, which employs a two stream model to separately generate the foreground and background, our model is a unified model and does not distinct foreground from backgrounds. Moreover, for each of the generated image, the boundary of the foreground subject is sharp and looks natural in the scene. Qualitatively, this is better than the results in~\cite{DBLP:conf/nips/VondrickPT16}, where the people in the scene are often blobs.
2) The model successfully generates motions patterns with high quality. We find that the generated motion sequences are highly recognizable for humans. Moreover, the skeletons extracted from the generated sequence are very similar from ground-truth skeletons. We make two remarks here. On one hand, this demonstrates that the generated pose is close to the objective. On the other hand, it in turn demonstrates that our model has successfully generated humans that can be recognized by the pose detector.
3) The identity of the appearance reference image is well preserved. For example, in Figure~\ref{fig:subfig:a} and Figure~\ref{fig:subfig:c}, the appearance reference image of the subjects are with white and black clothes respectively.
We can observe that the generated sequence share similar appearances with their reference images, and the appearance of the person is consistent in the generated sequence.

\begin{figure*}
\centering
\setlength{\tabcolsep}{0pt}
\renewcommand{\arraystretch}{0}

\subfigure[Walking.]{
\label{fig:36m:a}
\includegraphics[width=\textwidth]{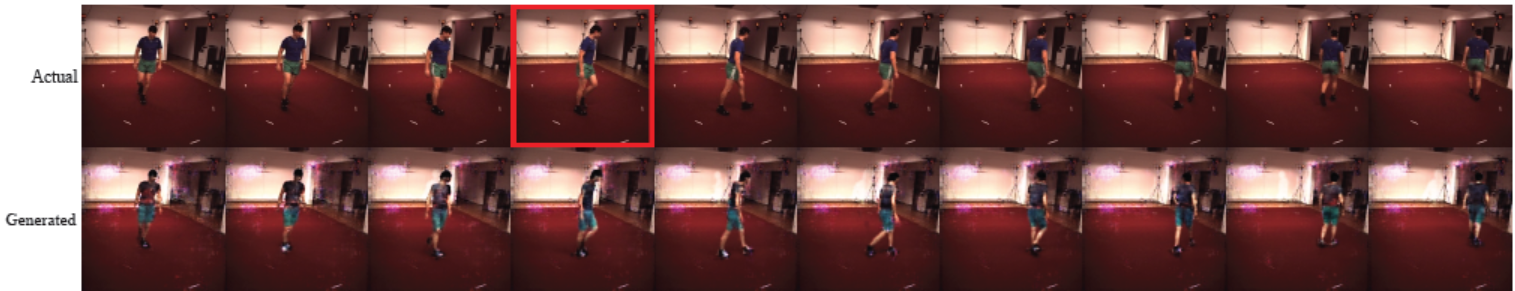}
}

\subfigure[Walking without background.]{
\label{fig:36m:b}
\includegraphics[width=\textwidth]{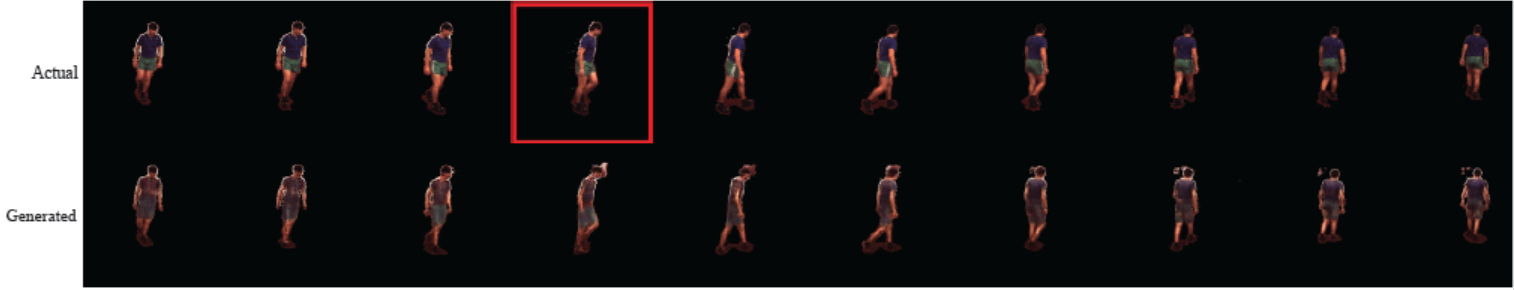}
}

\caption{Generation results on Human3.6M dataset. (a) Walking sequence with background. (b) The same sequence without background.}
\label{fig:human}
\end{figure*}

The generation results on Human3.6M dataset are visualized in Figure~\ref{fig:human}. Note that different from KTH, Human3.6M dataset contains color videos which are much more difficult to encode the appearance into latent space. We can observe that the appearance of generated sequence is close to the ground truth. The results demonstrate that the generator effectively transforms the color space into latent space, and finds good representation for the relationship between different body parts and their corresponding color. And we also present the results on Human3.6M without background in Figure~\ref{fig:36m:b}. We can observe that the network indeed transforms the appearance from the target image to generated sequence, without referring to the background information.

Overall, the proposed method shows promising results in generating human motions.
For more examples, please refer to our supplementary materials.

\subsection{Component Analysis}
\begin{figure}
\includegraphics[width=3.5in]{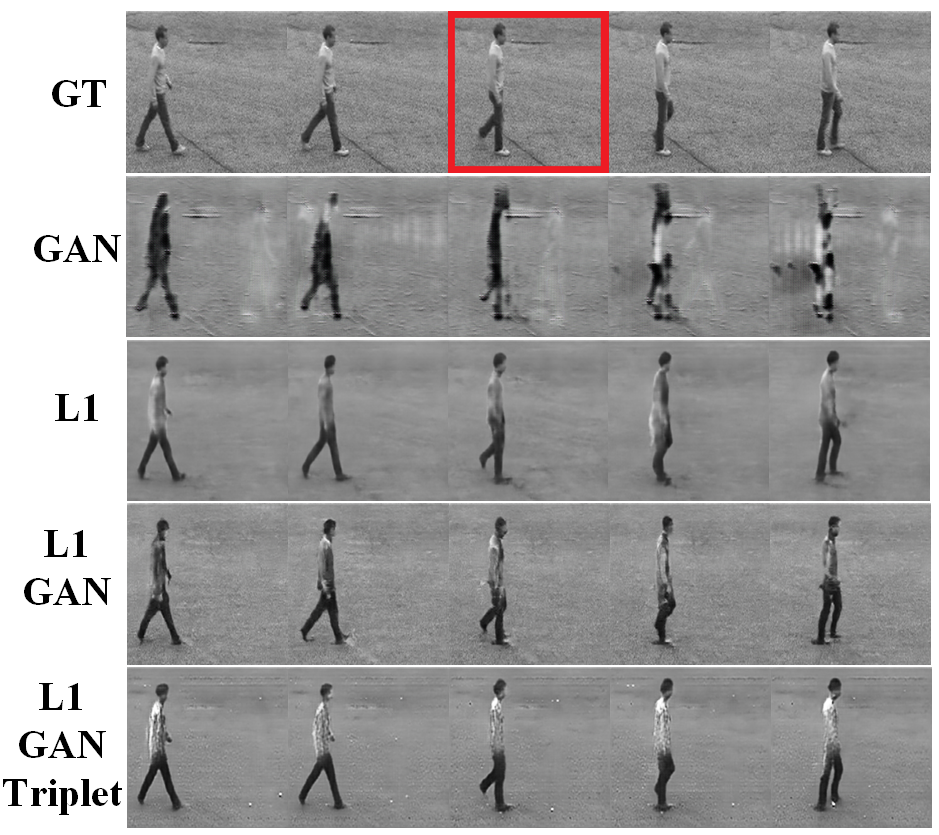}
\caption{Examples generated with different loss terms.}
\label{fig:loss}
\end{figure}

In this part, we study how the loss terms in Equation~\ref{eq:l} influence the generation results. The GAN loss and the L1 loss have been analyzed in~\cite{DBLP:journals/corr/IsolaZZE16} for image-to-image translation task, which demonstrates that using GAN loss alone will generate artifacts, and using L1 loss alone tends to generate the color averaged over the training set. The results in our experiments are illustrated in Figure~\ref{fig:loss}.
We can observe that the L1 loss and GAN loss have different effects for the generated videos. Specially, only using the GAN loss will result in severe artifacts in the video, some parts of the person are missing and color of the person keeps changing in the video. While only using L1 loss will degrade the generalization ability of the network. The third row of Figure~\ref{fig:loss} shows these results. No matter how we change the reference image, all the generated sequences have almost the same appearance, i.e., the average color of all the training samples. These adversary effects in these two kinds of losses are critical for training the generator. Combing L1 loss and GAN loss will generate better results, but the color is not consistent along the frames. Furthermore, introducing the triplet loss described in Section3.1 will stabilize the performance of generator. As shown in bottom two rows of Figure~\ref{fig:loss}, the generated sequence in bottom has consistent clothing color compared to the results which lack the help of triplet loss.

We also perform quantitative evaluations over the loss terms. We train an action classification framework~\cite{DBLP:journals/pami/DonahueHRVGSD17} on the KTH dataset, and then use it to classify the generated sequences. If the generated sequences are classified into the right action type, it demonstrates that the generated sequences can be recognized by the off-the-shelf classifier. The results are shown in Table~\ref{table:accuracy}.
As KTH dataset is a relatively easy dataset for action recognition, the trained classifiers achieve very good performance ($\sim$100\%) on the ground-truth test sequences. We observe that only using L1 or GAN loss leads to great performance drop, and combining the two loss terms significantly improves the recognition accuracy. Further adding triplet loss does not bring in further improvements, because the motion patterns are already clear enough for the classifier.

\begin{table}[]
\centering
\caption{Recognition accuracies (\%) on the generated sequences. GT denotes the ground-truth results. L1 denotes L1 loss, G denotes GAN loss, T denotes triplet loss.}
\label{table:accuracy}
\begin{tabular}{c|c|c|c|c|c}
\hline \hline
Action      & GT  & L1 & G  & L1+G & L1+G+T \\ \hline
Walking     & 100 & 83.1 & 43.1 & 93.8   & 93.8     \\
Running     & 98.5  & 76.9 & 41.5 & 92.3   & 92.3     \\
Hand waving & 100 & 95.4 & 47.7 & 100  & 100    \\ \hline \hline
\end{tabular}
\end{table}

\section{Conclusion}
In this work, we propose a skeleton-aided articulated motion generation method. In contrast to existing video generation methods, which usually lack geometric constraints for the foreground object, our method utilizes skeleton information as a guidance for the geometric change during the motion.
Experimental results show that by giving an appearance reference image and the skeleton sequence, our model can produce high-quality video sequences that not only preserve the appearance of the reference image, but also have clear motion pattern as the skeletons. The generated motion sequences are also recognizable by the off-the-shelf action recognition framework.


\bibliographystyle{IEEEbib}
\bibliography{mybib}

\end{document}